# A Diagnosis and Treatment of Liver Diseases: Integrating Batch Processing, Rule-Based Event Detection and Explainable Artificial Intelligence

*Ritesh Chandra\*,  Sadhana Tiwari,  Satyam Rastogi, and Sonali Agarwal*

Indian Institute of Information Technology Allahabad, Manipal Institute of Technology Jaipur, India

*rsi2022001@iiita.ac.in\*, rsi2018507@iiita.ac.in  satyamrastogi302@gmail.com, sonali@iiita.ac.in,*

**Abstract**

Liver diseases pose a significant global health burden, impacting many individuals and having substantial economic and social consequences. Rising liver problems are considered a fatal disease in many countries, such as Egypt and Moldova.  This study aims to develop a diagnosis and treatment model for liver disease using Basic Formal Ontology (BFO), Patient Clinical Data (PCD) ontology, and detection rules derived from a decision tree algorithm. For the development of the ontology, the National Viral Hepatitis Control Program (NVHCP) guidelines were used, which made the ontology more accurate and reliable. The Apache Jena framework uses batch processing to detect events based on these rules. Based on the event detected, queries can be directly processed using SPARQL. We convert these Decision Tree (DT) and medical guidelines-based rules into Semantic Web Rule Language (SWRL) to operationalize the ontology. Using this SWRL in the ontology to predict different types of liver disease with the help of the Pellet and Drools inference engines in Protege Tools, a total of 615 records were taken from different liver diseases. After inferring the rules, the result can be generated for the patient according to the rules, and other patient-related details, along with different precautionary suggestions, can be obtained based on these results.  These rules can make suggestions more accurate with the help of Explainable Artificial Intelligence (XAI) with open API-based suggestions. When the patient has prescribed a medical test, the model accommodates this result using optical character recognition (OCR), and the same process applies when the patient has prescribed a further medical suggestion according to the test report. These models combine to form a comprehensive Decision Support System (DSS) for the diagnosis of liver disease.

## 1. Introduction

Liver disease accounts for approximately 2 million deaths per year worldwide, 1 million due to complications of cirrhosis and 1 million due to viral hepatitis and hepatocellular carcinoma. Cirrhosis is the 11th top cause of death in the world, and liver cancer is the 16th. Together, it is responsible for 3.5% of all deaths globally [1]. It affects millions of lives and has a substantial economic and social impact. An appropriate diagnosis and effective management of liver diseases require in-depth knowledge of the underlying causes, signs and symptoms, risk factors, and proper medications. The combination and evaluation of numerous sources of information have become essential for comprehending complex illnesses because of the rapid increase in biological data.

At present, ontology has emerged as a formidable framework for the representation and organization of massive amounts of both structured and semi-structured data. In the current global context, there is a widespread shortage of doctors, with India[1] being particularly affected. Many struggle due to insufficient knowledge about appropriate medical care and regular health check-ups. To address this issue, we have developed a model that can be implemented as an application and website, especially in remote areas. This solution reduces patients' reliance on doctors and helps them to avoid unnecessary medical expenses.

A SWRL rules-based XAI makes precautionary suggestions more accessible and understandable for everyday users, presenting a novel approach. This is particularly beneficial for frontline workers like Accredited Social Health Activist (ASHA) workers and nurses, who need to comprehend the information. This clarity is crucial because not everyone in remote areas knows how to use mobile applications. When frontline workers conduct door-to-door campaigns, this approach can provide patients with instant guidance.

Ontology successfully uses a graph-based representation of knowledge to capture the links between various entities and their features simply and clearly. Knowledge graphs facilitate efficient searching, reasoning, and discovering new knowledge by displaying information graphically [2]. It facilitates the establishment of a universally accessible and standardized framework for information exchange between humans and machines. Eliminating ambiguities in terminology enables the efficient utilization of domain knowledge across various contexts. The phrase is initially described as a precise and formal specification of the terms within a certain field and their connections [3].

The foundational principle of the semantic web has evolved to a more advanced stage, leading to the development of many ontologies. The PCD ontology, when integrated with BFO [4], benefits from BFO's adherence to Open World Semantics and its reasoning capabilities, enhancing decision-making processes, particularly in liver disease management. This combination strengthens the structure and application of healthcare data, facilitating more informed clinical decisions.

A DSS is an information system that utilizes computer technology to integrate models and data to address unstructured or semi-structured problems. It facilitates various user interactions through a user-friendly interface. Both healthcare providers and patients alike can significantly utilize the proposed DSS. It not only aids healthcare professionals in diagnosing and treating medical conditions, but also enhances the provision of healthcare services from a distance, impacting the overall well-being and quality of life of patients [5].

---

[1] https://pmc.ncbi.nlm.nih.gov/articles/PMC11110446/

Application of the semantic web presents a viable solution for the dissemination and depiction of knowledge, hence enhancing an individual's proficiency. Ontology is considered a fundamental component of the semantic web. The term "adopted" refers to using a technology for knowledge representation within the context of its definition. The tool operates as a specialized lexicon, providing definitions for entities, attributes, and their interconnections [6].

This research work focuses on diagnosing liver-related diseases like cirrhosis and hepatitis C by developing a DSS that integrates BFO with the PCD ontology. DT rules are converted into RDF using RDFlib, with Apache Jena handling batch processing to detect events. SWRL rules and SPARQL queries are applied to make the ontology functional and to predict various liver diseases. The system also utilizes ChatGPT for enhanced recommendations and OCR to process medical test results, creating a comprehensive tool for liver disease management. The following points illuminate the main findings of this study:

- The collected data of different liver diseases is transformed into RDF triples to increase their quality and reusability for future usage.

- Conceptualizing the available knowledge of liver diseases via BFO and PCD ontologies to better understand liver diseases, precautions, treatment and create a knowledge graph.
- To develop a rule-based diagnosis system for liver patients, a decision tree is used, in which a set of rules can be defined with the help of classes, properties, and persons using SWRL.
- RDF data batch processing-based event detection using DT rules.

- The Knowledge-Driven Decision Support System enhances decision-making by incorporating factual information, established rules, and procedural guidelines.

- A user-level explanation model based on SWRL and XAI.

The rest of the paper is arranged as follows: The work that is currently available in the domain is covered in section II. Section III presents details of the methodology used, while section IV examines the experimental details and results. Section V and VI presents discussion, conclusion and future work respectively . Table 1 shows the abbreviations used in this manuscript.

Table 1: List of Abbreviations

| **Abbreviations** | **Name** | **Abbreviations** | **Name** |
| --- | --- | --- | --- |
| ALB | Albumin | CREA | Creatinine |
| ALP | Alkaline Phosphatase | GGT | Gamma-Glutamyl Transferase |
| ALT | Alanine Aminotransferase | PROT | Protein |
| AST | Aspartate Aminotransferase | SWRL | Semantic Web Rule Language |
| BIL | Bilirubin | DSS | Decision Support System |
| CHE | Cholinesterase | RDF | Resource Description Framework |
| CHOL | Cholesterol | DT | Decision Tree |
| BFO | Basic Formal Ontology | HCC | Hepatocellular Carcinoma |
| OWL | Ontology Web Language | OCR | Optical Character Recognition |
| XAI | Explainable Artificial Intelligence | NLP | Natural Language Processing |
| HCV | Hepatitis C Virus | | |

## 2. Related Work

Numerous studies have been proposed in recent times for detecting, preventing, and treating liver disease. These investigations have brought attention to the possible gaps in research and the interest in diagnosing diseases related to the liver.

Messaoudi et al. [7] introduced an ontological approach to liver cancer diagnosis, emphasizing medical imaging. Their method, including the LI-RADS method, aids in cancer detection, staging, and treatment, focusing on HCC. Banihashem et al. [8] devised a knowledge-based system using ontology and rules to detect fatty livers, employing a decision tree method. Based on 43 SWRL rules and the Drool inference engine, their system utilizes 70% clean data from 580 electronic medical records. AL-Marzoqi et al. [9] proposed a web-based methodology using ontologies to study liver viruses, aiming to assist medical professionals and students identify and diagnose viral hepatitis. Their evaluation involved querying the ontology for disease-related information, symptoms, signs, and laboratory findings.

Chan et al. propose an Electronic Health Records (EHR) based approach for analyzing clinical reports, particularly distinguishing HCC disorders. Their system employs Systematised Nomenclature of Medicine (SNOMED) ontology principles to retrieve and organize phrases associated with liver cancer, demonstrating efficacy through semantic queries and reasoning techniques with authentic patient data [10]. An ontology-based methodology is proposed for modeling patient cases, utilizing patients diagnosed with liver disease. A unique ontology called Liver Case Ontology (LICO) is presented to achieve these objectives, which incorporates established medical criteria for representing liver patient situations. This methodology sets the groundwork for a clinical experience-sharing platform for diagnosis, research, and education [11].

Chandra et al. developed an ontology for diagnosing and treating vector-borne diseases (VBDs), integrating data from Indian Health Records, medical mobile apps, and doctors' notes. Utilizing OCR and Natural Language Processing (NLP), they preprocess and extract entities from text data into RDF format using the PCD model. The ontology combines BFO, National Vector Borne Disease Control Programme (NVBDCP) guidelines, and RDF medical data with SWRL rules for diagnosis and treatment, aiding in developing decision support systems for VBDs [12].

Moawad et al. [13] developed the Ontology of Biomedical Reality (OBR) framework to construct a Viral Hepatitis Ontology covering the A, B, C, and D viruses, facilitating information sharing among intelligent systems and doctors. The Viral Hepatitis Ontology utilizes the Web Ontology Language (OWL), a language that has recently emerged as the standard for the semantic web. By creating the Viral Hepatitis Ontology, both intelligent systems and doctors will be able to share, think about, and use this information in a variety of ways [14]. Panigrahi et al. [15] created a Clinical Decision Support System (CDSS) for hepatitis B diagnosis in rural India, utilizing an expert system with 59 rules. Another work proposes a novel multi-label classification methodology using a Convolutional Neural Network (CNN) framework to enhance HCC lesion identification and tumor detail extraction through ontology utilization [16].

An ontology is developed to categorize hepatitis-related diseases, facilitating semantic search and retrieval [18, 19]. Some studies utilize medical imaging techniques like MRI or CT scans for diagnosis. Decision trees identify Nonalcoholic fatty liver disease (NAFLD) factors like Body Mass Index (BMI), Waist-to-Hip Ratio (WHR), Triglycerides (TG), Fasting Blood Glucose (GLU-AC), Systolic Blood Pressure (SP), and Serum Glutamate Pyruvate Transaminase (SGPT) with 75% accuracy [20]. Logistic regression predicts Alcoholic Fatty Liver Disease (AFLD) based on age, gender, BMI, SGPT, TG, and Total Cholesterol (TC) levels, achieving 61.4% accuracy [21]. AI-

Marzoqi et al. [22] focus on ontological approaches for hepatobiliary diseases, analyzing liver-based systems formation. Table 2 represents the summary of the other existing works.

Table 2: Summary of existing works

| Author | Objectives | Use of Ontology | Simulation | Tools and Languages |
|---|---|---|---|---|
| Yim et al. [23] | The application of reference resolution approach in the representation of semantically liver cancer tumors. | The process of identifying and analyzing the characteristics of liver cancer through textual reporting. | Precision and Recall measures. | WordNet, MetaMap and UMLS. |
| Kaur et al. [24] | The ontological approach is a method used in the detection of liver cancer. | Providing a semantic representation of the disease known as liver cancer. | The utilization of SPARQL to extract information about liver cancer disease. | SPARQL, OWL, and OntoGraph. |
| Gibaud et al. [25] | The OntoVIP framework is a method utilized for the annotation of medical images. | The utilization of pre-existing ontologies to achieve semantic representation. | The establishment of a shared lexicon. | RadLex and SPARQL. |
| Levy et al. [26] | The application of semantic reasoning in conjunction with image annotations for tumor evaluation. | Engaging in logical analysis of the resultant image and providing annotations for evaluating tumor lesions. | Converting AIM Data into OWL Format. | OWL, XML, SQWRL, and SWRL. |

After reviewing numerous scholarly articles in the relevant literature, it is observed that no one has discussed the proper upper-level ontology for developing a liver disease ontology. This work uses BFO to develop ontologies and extract SWRL rules based on DT rules. The model presents a substantial advancement over the work of Banihashem et al. [8]. While their study focuses solely on fatty liver identification using SWRL rules and ontology, this approach expands to cover a broader range of liver-related diseases. This ontology also includes all medical guidelines related to liver disease. Previous works do not discuss complete knowledge on one platform.

The proposed work defines rules-based disease prediction and generates precautionary suggestions using XAI and events based on DT rules for batch processing using the Apache Jena framework. Based on these events, queries can be processed easily through SPARQL. Until now, no one has discussed an event-based model for batch processing-based event detection for liver disease, which reflects the uniqueness of the proposed work. The model is also integrated with OCR and NLP techniques and PCD ontology, making it comprehensive because it predicts diseases and suggests treatments based on further medical test reports, which are explained in the subsequent section

## 3. Methodology

This section provides a detailed overview of the methodology applied in this research, including the approaches, techniques, and frameworks to address the research problem effectively.

### 3.1 Architecture of the suggested model

Figure 1 represents the model architecture of the proposed work which includes various phases such as data pre-processing, conversion of preprocessed data into RDF format, extracting rules from the decision tree, batch processing through Apache Jena, query processing, complex event detection, ontology development, text extract through OCR and ChatGPT based explanation.

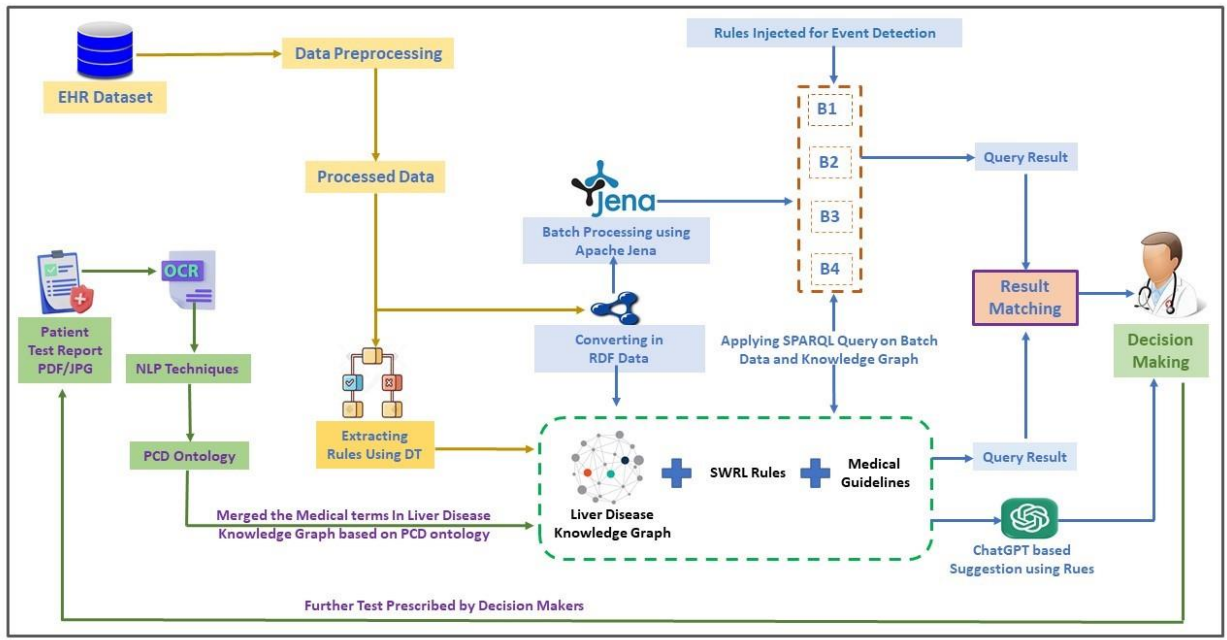

Figure 1: Structure of the proposed model

### 3.2 Dataset Description

The data is obtained from the UCI Machine Learning Repository[2], which is a CSV dataset that contains 615 entries and 14 attributes, including Patient ID, Sex, Age, and Category in which the patient falls, like healthy, a hepatitis C patient, a fibrosis patient, a cirrhosis patient, or a patient who does not have any diseases but just shows symptoms. It also has laboratory data like ALB, ALP, ALT, AST, BIL, CHE, CHOL, CREA, GGT, and PROT. To develop the ontology, we used medical guidelines based on the National Viral Hepatitis Control Program (NVHCP)[3] to ensure accuracy and comprehensive domain knowledge coverage.

### 3.3 Dataset Preprocessing

The dataset must be preprocessed, as it differs from what is required. Initially, we fill the missing values in the dataset with the mean of the corresponding column using the 'fillna' function. We then mapped non-numeric values to numeric values for specific columns. For example, the 'Category' column is mapped using a dictionary that assigns numeric values to different categories, i.e., 0=blood donor, to 0 meaning healthy; 0s=suspect blood donor, to 1 which means a person having symptoms of liver disease but not having a liver disease; 1=hepatitis, to 2 meaning a hepatitis patient; 2=fibrosis, to 3 which means a fibrosis patient; and 3=cirrhosis, to 4 which means a cirrhosis patient.

Similarly, a dictionary maps the 'Sex' column, assigning numeric values to different genders (0 for females and 1 for males). We add another column to store the UIDs of the patients. This preprocessing of the dataset transformed the categorical and non-numeric values into numeric representations, enabling further analysis and processing of the data.

### 3.4 RDF Conversion

The World Wide Web Consortium (W3C) collaborated in the creation and standardization of RDF, a standard for defining online resources and data transmission. It is a data representation technique that establishes connections between data elements. RDF allows users to successfully integrate data from different sources by decoupling it from its format. This makes it possible to use, integrate, query, and update numerous schemas as a single entity without changing the data instances. In this work, we convert the CSV dataset into RDF using Python by applying the following steps:

1. To convert the dataset into RDF data format, the relevant libraries for CSV parsing and RDF manipulation, such as 'csv' and 'rdflib', must be imported.

---

[2] https://archive.ics.uci.edu/ml/datasets/HCV+data
[3] https://nvhcp.mohfw.gov.in/about_us

2. Create RDF namespaces using 'Namespace' from 'rdflib' to represent the RDF. Namespaces provide unique and globally recognized identifiers for terms used in RDF data.
3. Initialized an RDF graph object using 'Graph' from 'rdflib' to store the RDF data.
4. Open the CSV file that contains the dataset and use 'csv.dictreader' to read the CSV file, treating each row as a dictionary with column names as keys.
5. Iterate over each row in the CSV data, and for each row, the values for different columns are extracted and assigned to variables.
6. RDF triples are created in which the subject is chosen as the UID of patients; this is done via 'URIRef' from 'rdflib'. The type of subject is also added using the RDF "type" property; this is treated as MedicalRecord.
7. Adding RDF triples for each attribute in the row using the 'graph.add' method and defining predicates and object values via `URIRef` from the schema.org' namespace and `Literal` from `rdflib`. Specify the appropriate data type (e.g., XSD.integer, XSD.float) for each literal value using the `datatype` parameter.
8. Serialized the RDF graph data using the 'serialize' method of the graph object and specified the output format as 'xml'.

### 3.5 Rule Extraction Using Decision Tree

Using a decision tree to get rules from a trained model is called rule extraction. Each line from the root node to a leaf node represents a different rule as shown in Figure 3. These rules are made up of conditions (based on feature levels) that cause certain results or predictions to happen. To turn the tree's structure into a style that humans can understand, the process usually uses functions in Python's scikit-learn library, such as export_text() steps are shown in Table 3 in the form of an algorithm. This method gives clear information for making decisions, which makes it useful in areas that need to be clear, like healthcare tests and compliance-driven areas. Table 4 shows the rules extracted from the dataset using DT. Figure 2 shows the feature importance values across different features.

Table 3: Algorithm:for building, extracting, and converting DT rules to SWRL Rules

| |
|---|
| Input: Dataset with features X (independent variables) and target y (dependent variable). Output: Trained Decision Tree model, extracted decision rules, and SWRL rules. |
| START<br>Step 1: Import Libraries<br>   a. Import necessary libraries (e.g., pandas, sklearn.tree, DecisionTreeClassifier, export_text).<br>   b. Include additional libraries for SWRL conversion (e.g., rdflib for OWL integration). |

Step 2: Load and Prepare Data
   a. Load the dataset into a DataFrame.
   b. Identify relevant features (X) and the target variable (y) based on the problem statement.
   c. Handle missing or inconsistent values and normalize data if necessary.

Step 3: Perform Attribute Selection
   a. Remove irrelevant or redundant features to reduce dimensionality.
   b. Use a feature selection criteria like the Gini index for optimal performance.

Step 4: Initialize the Model
   a. Create an instance of DecisionTreeClassifier with desired parameters (e.g., `criteria='gini'`).

Step 5: Train the Model
   a. Fit the DecisionTreeClassifier to the dataset using `fit(X, y)`.

Step 6: Validate Model Performance
   a. Split the dataset into training and validation subsets.
   b. Use cross-validation (e.g., `cross_val_score`) to evaluate model reliability.

Step 7: Extract Decision Rules
   a. Use the `export_text()` function to extract rules from the trained decision tree.

Step 8: Convert Rules to SWRL
   a. For each rule:
    - Map each condition to the antecedent (IF part) in SWRL.
    - Map the corresponding outcome to the consequent (THEN part) in SWRL.
   b. Handle complex conditions (e.g., logical operators like AND, OR) using SWRL syntax.

Step 9: Save SWRL Rules
   a. Save the rules in a format compatible with ontology systems (e.g., OWL files) using libraries like rdflib.
END

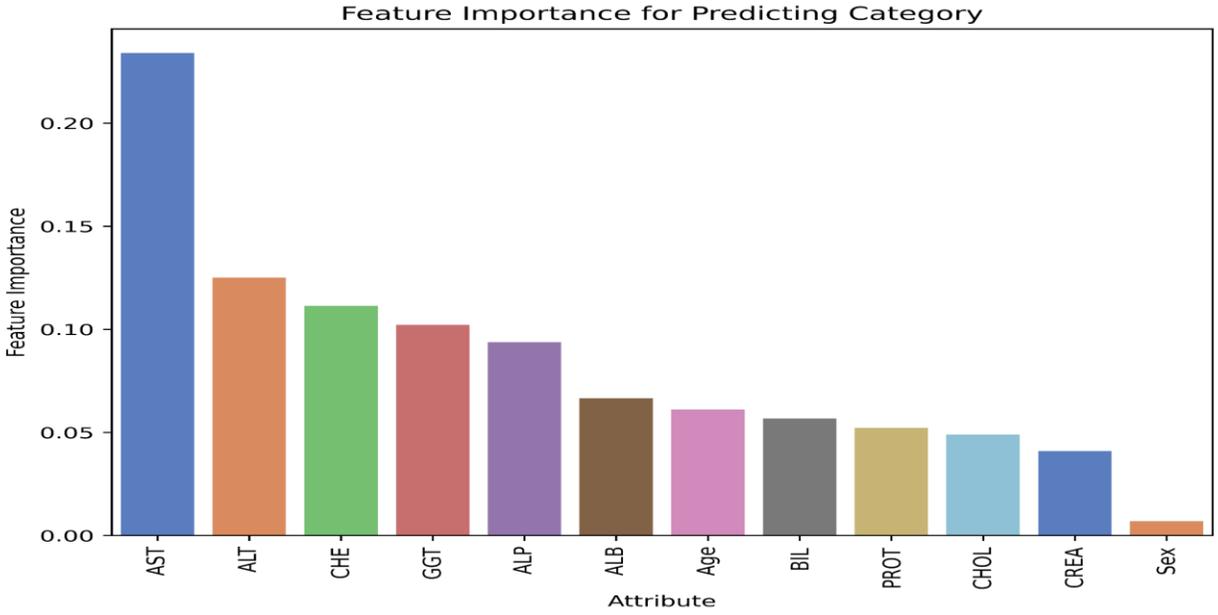

Figure 2: Feature Importance value across different features

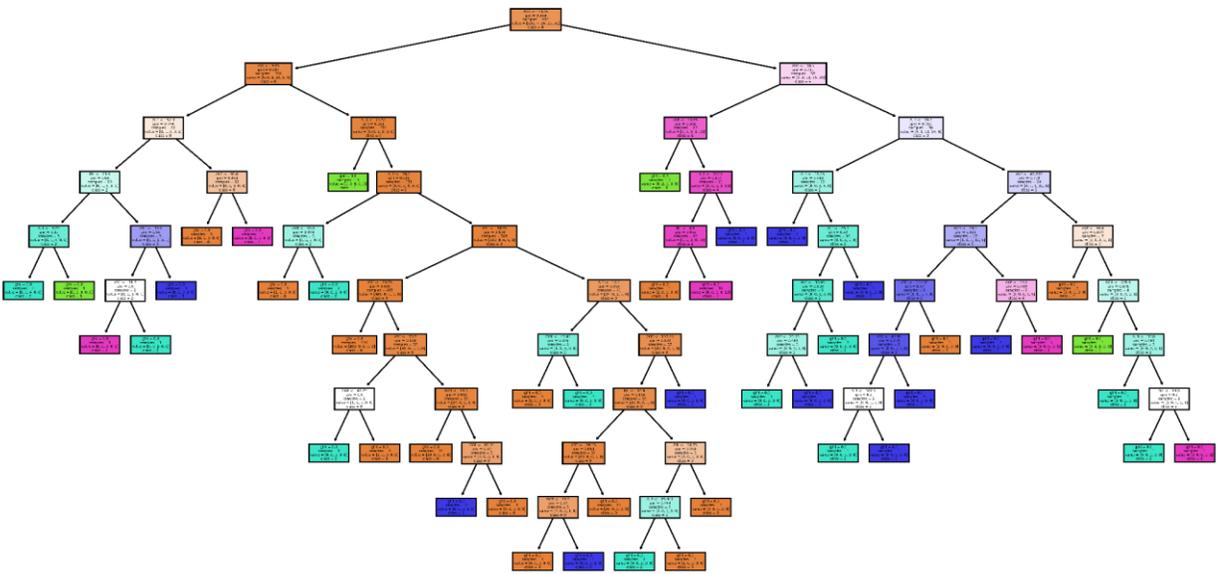

Figure 3: DT rules which are extracted from the dataset

Table 4: Represents the DT rules which are extracted from the dataset

| Serial Number | If Clause | Then Statement |
|---|---|---|
| 1 | If AST <= 53.05 and ALP > 52.3 and ALT <= 9.65 and ALP <= 98.6 | Patient is Healthy |
| 2 | If AST <= 53.05 and ALT > 9.65 and ALB <= 25.55 | Patients show some potential signs of liver abnormalities or diseases(though they don't meet the criteria for a specific condition). |
| 3 | If AST <= 53.05 and ALP <= 52.3 and BIL <= 11.0 and ALT <= 9.25 | Patient has Hepatitis C |
| 4 | If AST <= 53.05 and ALT > 9.65 and ALB > 25.55 and ALP > 28.2 and AST <= 38.05 and AST > 33.05 and AST > 33.2 and GGT > 83.3 and GGT <= 88.35 | Patient has Fibrosis. |
| 5 | If AST <= 53.05 and ALT <= 9.65 and ALP > 52.3 and ALP > 98.6 and BIL > 11.0 | Patient has Cirrhosis. |
| 7 | If AST <= 53.05 and ALB > 25.55 and ALT > 9.65 and ALP > 28.2 and AST <= 38.05 and AST <= 33.05 | Patient is Healthy. |
| 8 | If AST <= 53.05 and ALB > 25.55 and ALT > 9.65 and ALP <= 28.2 and ALB <= 43.8 | Patient is Healthy. |

The rules extracted in Table 4, based on DT, also meet the ground truth parameters, demonstrating that the rules are valid and reliable from the patient's perspective.

### 3.6 Batch Processing using Apache Jena

Batch processing is a technique that sequentially processes the data rather than processing the entire dataset at once. It involves dividing the data into smaller portions, chunks, or batches and processing them sequentially, which is useful when dealing with large or continuously streaming datasets. Apache Jena, an open-source Java framework, is used for batch processing. It provides comprehensive tools and libraries for working with RDF data, performing SPARQL queries, and building knowledge graphs. Apache Jena enables tasks such as parsing, storing, querying, and reasoning over RDF data. It supports various RDF syntaxes, provides APIs for RDF manipulation, and offers functionalities for batch streaming, ontology modeling, and RDF dataset management.

The parameters for batch streaming, such as the batch size (`batchSize`) and the delay between batches in seconds ('delaySeconds'), have been set. and an empty Jena model to hold the RDF data [27] is created. The RDF file needs to be read into the model using the file path and a base URI to resolve relative URIs. To store values for specific properties, a list is initialized, and to track the count of triples processed, a 'count' variable is defined. Table 5 shows the algorithm for the Batch Processing for RDF.

Table 5: Batch Processing Algorithm for RDF Data

| **Algorithm for the Batch Processing for RDF** |
|---|
| Iterate over the triples and process each triple based on its predicate and object value:<br>● Print the triple for observation or further processing.<br>● Check if the predicate corresponds to the laboratory properties (e.g., ALT, AST, GGT, etc.) and add their respective values to the appropriate lists.<br>● Increment the count.<br>● Check if the batch size is reached (`count % batchSize == 0`).<br>    ○ If the batch size is reached:<br>        ■ Perform batch processing if all lists have the same size (`processBatch()`).<br>        ■ Clear the lists to release memory.<br>        ■ Sleep for the specified delay between batches.<br>Process any remaining triples that do not form a complete batch. |

### 3.6.1 Event Detection based on Decision Tree Rules

An event can be detected on the batch-processed data by applying some rules, conditions, or filters. Event detection is discovering and recognizing certain occurrences or patterns of interest within a data stream [28]. On the data that satisfies the rules, apply SPARQL queries to get the relevant information. Create a SPARQL query string to select the required variables (e.g., SNo, ALT, AST, GGT) and apply any desired filtering or conditions.

To implement this, the `QueryExecutionFactory.create()` method creates a query execution object with the query string and the Jena model. Execute the query using `execSelect()` to obtain the result set, iterate over the result set, and retrieve the values for each variable that is needed according to the query (for example, SNo, ALT, AST, GGT, PROT, BIL, ALP, etc.), and then process and print the query results. Figure 4. shows the batch processing and rules injected on batches.

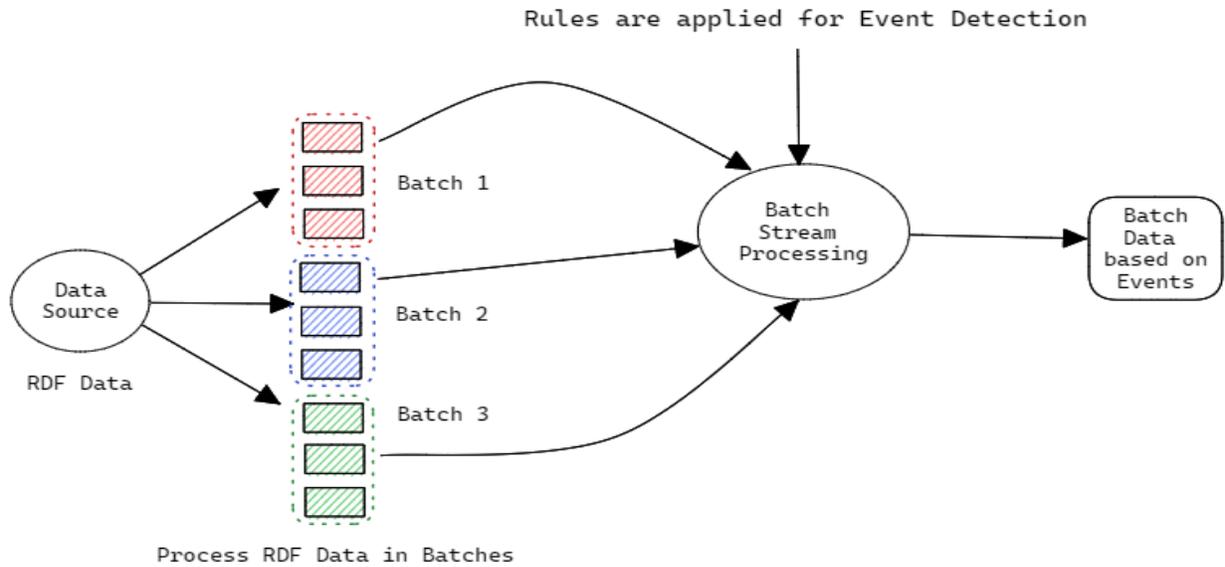

Figure 4: Batch processing and event detection based on DT rules

### 3.7 PCD Ontology

The PCD ontology is designed to describe the EHR clinical data components, encompassing clinical principles related to healthcare activities that occur during a patient's visit. When OCR [12] extracts the information from a patient's test report of a patient and processes textual data through NLP techniques, the relevant information is extracted based on the specified domain using techniques such as Word2Vec or FastText. The extracted words are mapped through the PCD ontology and converted into an RDF database frame view[29][30]. This process is illustrated in Figure 5. Subsequent steps are managed according to the proposed Liver Disease Ontology, as depicted in Figure 1.

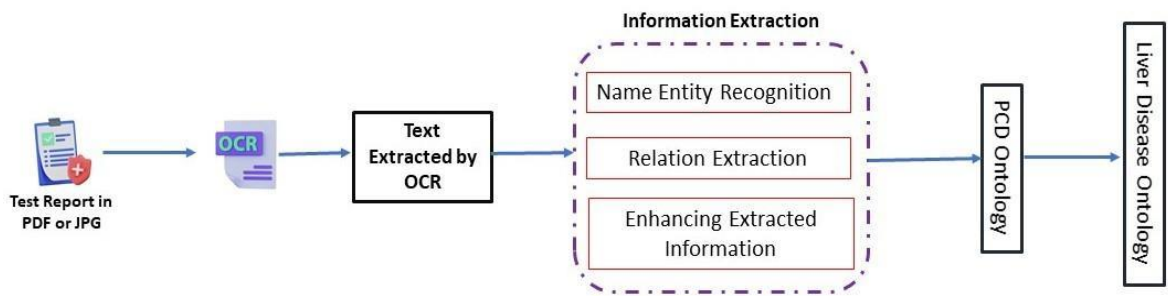

Figure 5: Information extraction method for PCD Ontology

### 3.8 Liver Diseases Ontology development using BFO

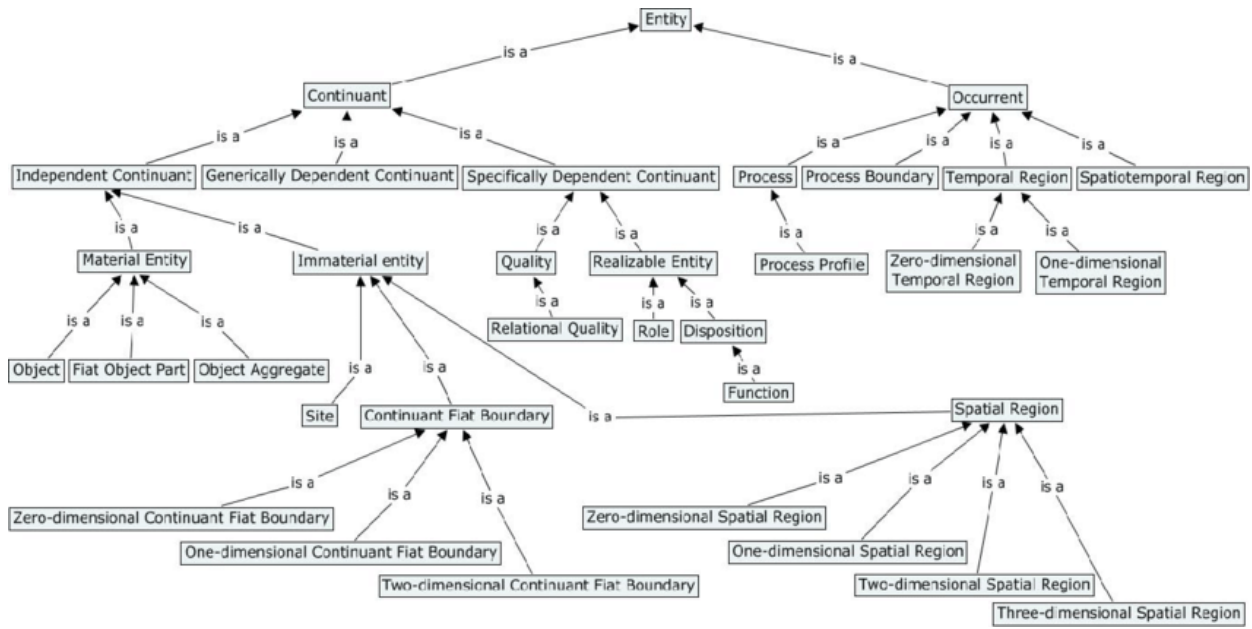

Figure 6: BFO[4] Hierarchy (Version 2.0) [12]

Using the RDF data, a knowledge graph has been constructed following the concepts and principles provided by the BFO. The BFO is an upper-level ontology designed to support information retrieval, analysis, and integration in scientific and other domains. It is a top-level ontology and a foundational framework for organizing and representing knowledge. A knowledge graph is a structured representation of knowledge and data. This knowledge is represented via classes, attributes, properties, and individuals. Entities could be events, situations, or concepts, and they are specified with a formal semantics that allow them to be processed by humans and machines. It is often maintained and visualized as a graph structure in a graph database.

BFO divides an entity into two types: continuous classes and occurring classes. Continuants persist through time and are characterized by maintaining their identity despite changes in their properties or attributes. Continuants are then subdivided into three classes: independent, genetically dependent, and specifically dependent. whereas occurrences are entities that occur, happen, or develop over time. They are also subdivided into four classes: process, process boundary, temporal region, and spatiotemporal region, as shown in Figure 6. This ontology or knowledge graph is developed using the Protégé 5.5 tool.

### 3.8.1 Classes

Classes represent categories or types of entities in a knowledge graph. It defines the common characteristics that instances within a class share. In BFO, an entity is divided into Continuant or

---

[4] https://basic-formal-ontology.org/

Occurrent classes. Continuant classes are further subdivided into three classes, shown in Figure 7.

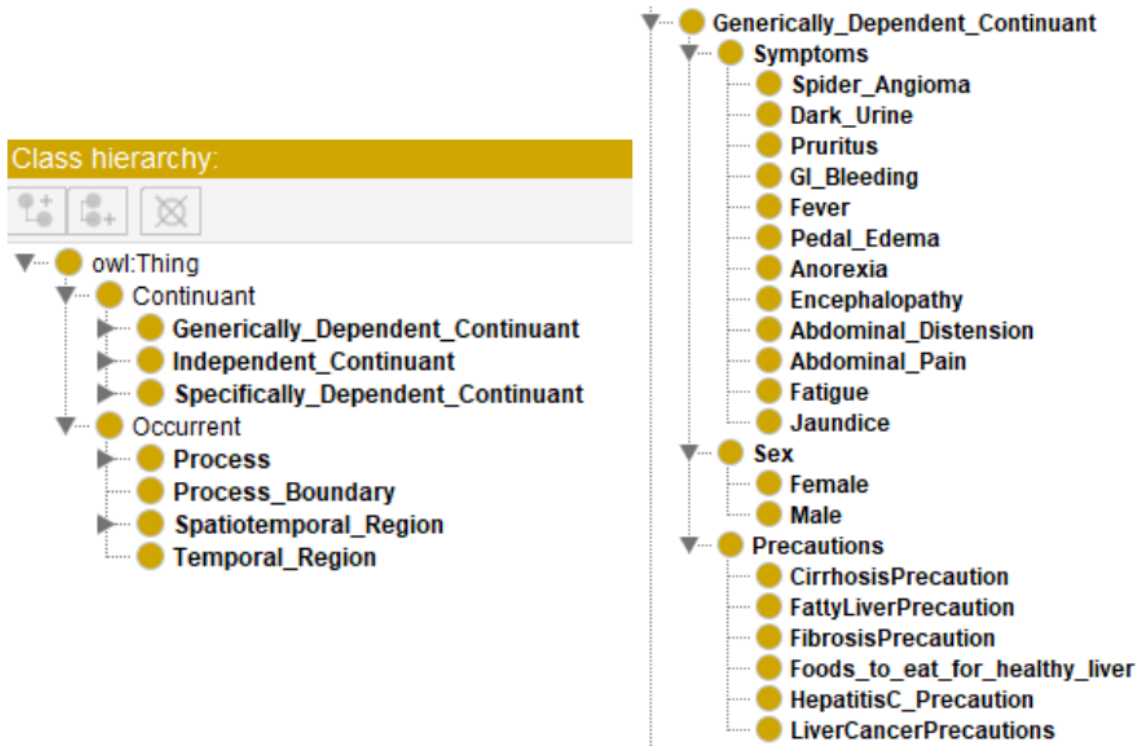

Figure 7: Class Hierarchy and Generically Dependent Continuant

- Independent Continuant: An entity that possesses its own identity independent of other entities.
- Generically Dependent Continuant: These entities depend on other entities in a generic or general sense.
- Specifically Dependent Continuant: (A specifically dependent continuant is a dependent continuant that exists only as a part of a particular independent continuant, meaning it is a dependent continuant that is part of a particular independent continuant.) or (Specifically dependent entities are entities whose existence or identity depends on the existence of other entities.) They rely on or depend on other entities for their status.

Generically Dependent Continuant has 3 subclasses, which are shown in figure 7:
- Symptoms: Symptoms of liver diseases are generically dependent continuants as they depend on the presence of specific diseases or medical conditions to be evident.
- Sex: Sex is a generically dependent continuant as it relies on the presence of individuals (in this case, patients) to have specific sex attributes.
- Precautions: Precautions for Diseases are preventive measures typically generically dependent continuants. They depend on the presence of specific diseases but can apply to multiple instances of those diseases.

Independent Continuant has five subclasses, which are shown in Figures 8 and 9, namely:

- Healthcare Providers: This class contains different types of people involved in patient care, such as pharmacists, nurses, and doctors, who have different kinds of doctors who assess or operate on liver diseases. Healthcare Providers are individual entities with their own identities and do not depend on other entities for their existence. Thus, they belong to the Independent Continuant.
- Hospitals are independent entities with their own identity and do not rely on other entities. Hence, they are also Independent Continuants.
- Pathology Labs: This contains different types of diagnostic tests. Pathology labs are standalone entities that do not depend on other entities and are also Independent Continuants.
- Risk Factors: Risk factors such as genetic predispositions or environmental factors can exist independently, making them independent continuants.
- Liver Diseases: As disease entities, liver diseases are independent continuants with their own identities.

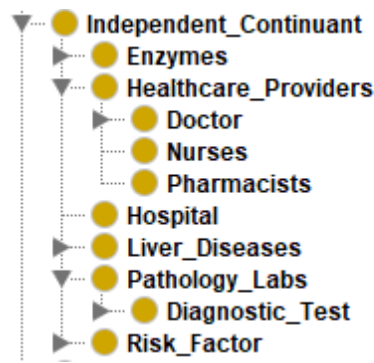

Figure 8: Independent Continuant

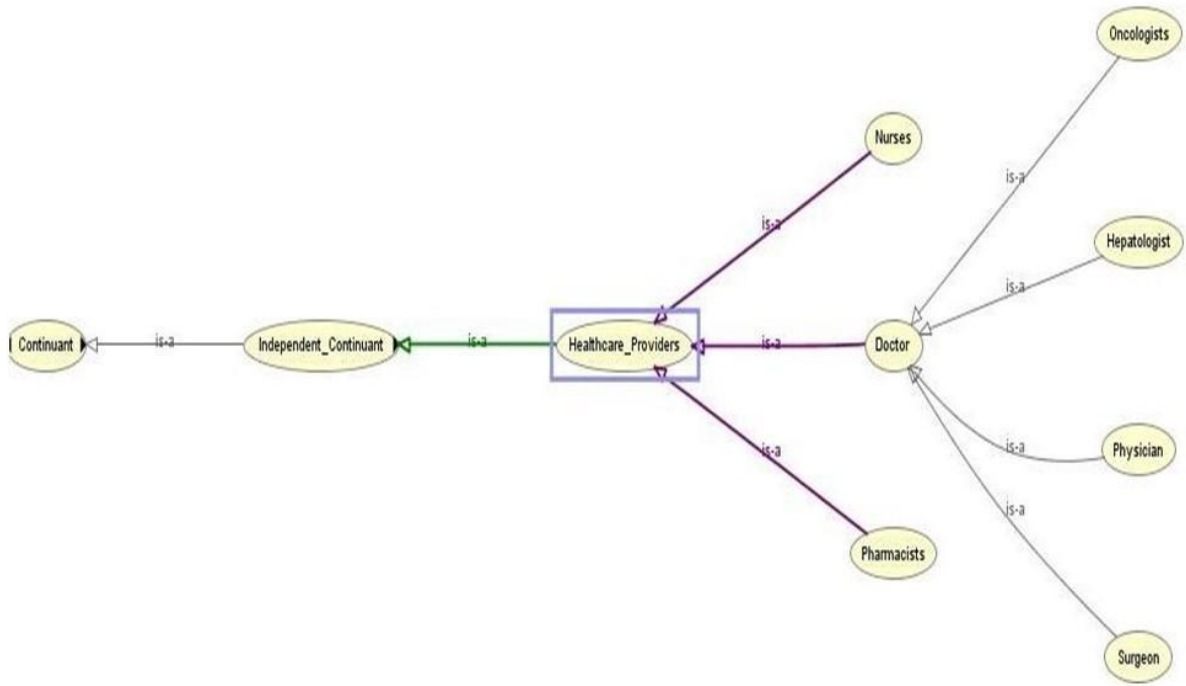

Figure 9: Classes under Healthcare_Providers using OWLVIZ

Specifically, Dependent Continuant has four subclasses shown in Figure 10.

- Allergies: Allergies depend on specific individuals (patients) for their existence.
- Treatments: Treatment for diseases is a specifically dependent continuant. It depends on the existence of individual diseases, and different treatments are specific to particular diseases.
- Category: The disease category can be explicitly considered dependent as it relies on individual diseases to be classified into specific categories.
- Patient: This contains subclasses for Patients like Personal Details and Medical History. It is specifically dependent because each patient is a unique individual with their own medical history, attributes, and records.

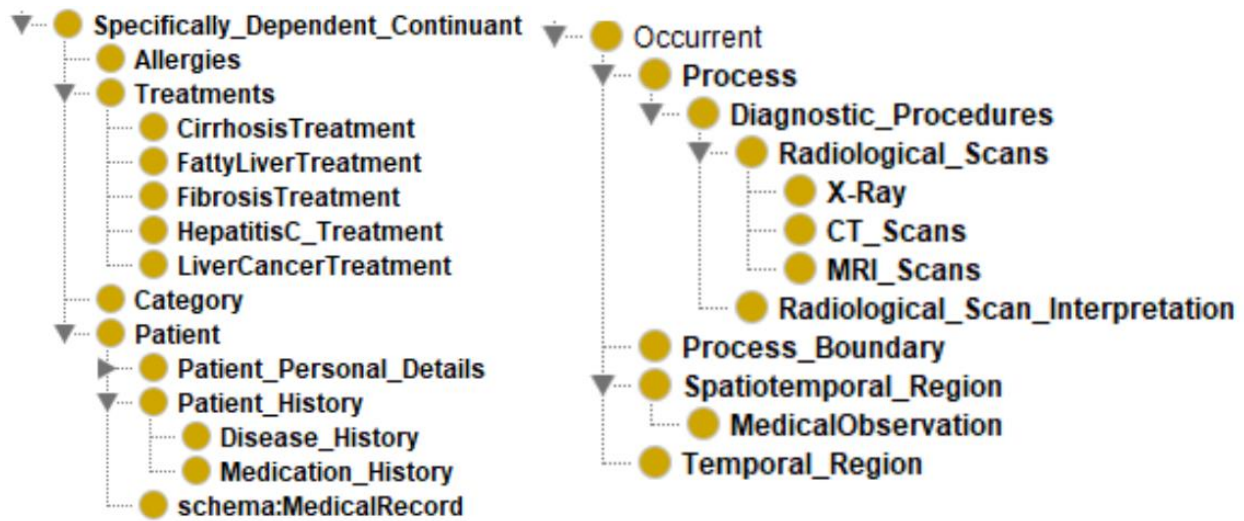

Figure 10: Specifically Dependent Continuant and Occurrent classes

Occurrent classes, like Continuant classes, are also further subdivided. They are subdivided into four classes: Process, Process Boundary, Spatiotemporal Region, and Temporal Region. The Process subclass represents an ongoing or continuous series of actions or activities. In the present case, the Process subclass has one, 'Diagnostic Procedure', which contains Radiological Scans like X-rays, CT Scans, and MRI Scans and their Interpretations. Spatiotemporal region refers to an area or space in which events, processes, or occurrences take place over a specific time; therefore, it has a subclass called 'MedicalObservation', which would contain Medical Observation of the Patient over time.

### 3.8.2 Properties

Properties in a knowledge graph are used to describe relationships between entities (individuals) and provide some additional information about those entities. There are two main properties of a knowledge graph:

1.) Object Properties: It establishes relationships between entities in the graph, i.e., it links one entity to another entity or a class, indicating a connection between them shown in Table 6.
2.) Data Properties: It assigns specific data values to entities shown in Table 7.

Table 6: Object Properties of the Ontology

| Properties | Explanation |
| --- | --- |

| has_Symptom | This Property denotes the symptoms that the patient is experiencing. |
|:---:|:---|
| is_SymptomOf | This Property denotes the correlation between the symptoms and their corresponding disease. |
| hasHealthInsurance | This Property denotes whether a Patient has Health Insurance. |
| hasPrecautions | This Property correlates the Precautions to the Liver_Diseases |
| is_Alcoholic | This denotes whether the Patient is Alcoholic or not. |

Table 7: Data Properties of the Ontology

| Properties | Explanation |
|:---:|:---:|
| hasValueALB | This Denotes the Value of Albumin Enzymes in IU/L. |
| hasValueALP | This Denotes the Value of Alkaline Phosphatase Enzymes in IU/L. |
| hasValueALT | This Denotes the Value of Alanine Transaminase Enzymes in IU/L. |
| isCirrhosisPatient | This Indicates whether the patient has cirrhosis or not. |
| isHealthy | This Denotes whether the Patient is Healthy or not. |
| Sex | This Property denotes the gender of the Patient. |

### 3.9 Conversion of Decision Tree Rules to SWRL Rules

An SWRL language introduces inference rules for extracting implicit knowledge from ontology. The knowledge models are expressed through Web Ontology Language (OWL). The SWRL is recognized as a prominent framework for expressing knowledge in the form of rules.The rules establish a logical relationship between a preceding condition (body) and a resulting condition (head). 36 rules are developed from DT, and 21 are developed through medical guidelines certified by government authorities[5]. Some converted DT rules to SWRL rules are shown in Table 8.

Table 8: Some Decision Tree Rules with their corresponding SWRL Rules

| S No. | DT Rules | SWRL Rules |
|---|---|---|
| 1 | If AST <= 53.05 and ALP <= 52.3 and BIL <= 11.0 and ALT <= 9.25 then Patient has Hepatitis C. | Patient(?x) ^ hasValueAST(?x, ?ast) ^ swrlb:lessThanOrEqual(?ast, "53.05"^^xsd:float) ^ hasValueALP(?x, ?alp) ^ swrlb:lessThanOrEqual(?alp, "52.3"^^xsd:float) ^ hasValueBIL(?x, ?bil) ^ swrlb:lessThanOrEqual(?bil, "11.0"^^xsd:float) ^ hasValueALT(?x, ?alt) ^ swrlb:lessThanOrEqual(?alt, "9.25"^^xsd:float) -> isHepatitisCpatient(?x, true) |
| 2 | If AST <= 53.05 and ALT <= 9.65 and BIL <= 11.0 and ALT > 9.25 and ALP <= 52.3 then the patient shows some potential signs of liver abnormalities or diseases (though they don't meet the criteria for a specific condition). | Patient(?x) ^ hasValueAST(?x, ?ast) ^ swrlb:lessThanOrEqual(?ast, "53.05"^^xsd:float) ^ hasValueALT(?x, ?alt) ^ swrlb:lessThanOrEqual(?alt, "9.65"^^xsd:float) ^ hasValueBIL(?x, ?bil) ^ swrlb:lessThanOrEqual(?bil, "11.0"^^xsd:float) ^ swrlb:greaterThan(?alt, "9.25"^^xsd:float) ^ hasValueALP(?x, ?alp) ^ swrlb:lessThanOrEqual(?alp, "52.3"^^xsd:float) -> isShowingSigns(?x, true) |
| 3 | If AST <= 53.05 and ALP <= 52.3 and BIL > 11.0 and AST <= 33.9 and AST <= 31.2 and ALT <= 9.65 then the patient has Cirrhosis. | Patient(?x) ^ hasValueALP(?x, ?alp) ^ swrlb:lessThanOrEqual(?alp, "52.3"^^xsd:float) ^ hasValueBIL(?x, ?bil) ^ swrlb:greaterThan(?bil, "11.0"^^xsd:float) ^ hasValueAST(?x, ?ast) ^ swrlb:lessThanOrEqual(?ast, "31.2"^^xsd:float) ^ hasValueALT(?x, ?alt) ^ swrlb:lessThanOrEqual(?alt, "9.65"^^xsd:float) -> isCirrhosisPatient(?x, true) |
| 4 | If AST <= 53.05 and ALT <= 9.65 and ALP <= 52.3 and AST > 33.9 and BIL > 11.0 then the patient | Patient(?x) ^ hasValueAST(?x, ?ast) ^ swrlb:lessThanOrEqual(?ast, "53.05"^^xsd:float) ^ hasValueALT(?x, ?alt) ^ |

---

[5] https://main.mohfw.gov.in/sites/default/files/Technical%20and%20Operational%20_LOW%20RISE_0.pdf

| | has Fibrosis. | swrlb:lessThanOrEqual(?alt, "9.65"^^xsd:float) ^ hasValueALP(?x, ?alp) ^ swrlb:lessThanOrEqual(?alp, "52.3"^^xsd:float) ^ swrlb:greaterThan(?ast, "33.9"^^xsd:float) ^ hasValueBIL(?x, ?bil) ^ swrlb:greaterThan(?bil, "11.0"^^xsd:float) -> isFibrosisPatient(?x, true) |
|---|---|---|
| 5 | If AST <= 53.05 and ALP > 52.3 and ALT <= 9.65 and ALP <= 98.6 then the patient is Healthy | Patient(?x) ^ hasValueAST(?x, ?ast) ^ swrlb:lessThanOrEqual(?ast, "53.05"^^xsd:float) ^ hasValueALP(?x, ?alp) ^ swrlb:greaterThan(?alp, "52.3"^^xsd:float) ^ swrlb:lessThanOrEqual(?alp, "98.6"^^xsd:float) ^ hasValueALT(?x, ?alt) ^ swrlb:lessThanOrEqual(?alt, "9.65"^^xsd:float) -> isHealthy(?x, true) |

### 3.10 Rules are extracted based on medical guidelines

To make the ontology more reliable and efficient, we generate rules based on ground truth for liver-related diseases, as per NVHCP. Table 9 shows some of SWRL rules for liver-related diseases based on NVHCP guidelines.

Table 9: Rules based on medical guidelines for Hepatitis C and Chrissosis

| Serial Number | Rule for | Condition | Action | SWRL Rule |
|---|---|---|---|---|
| 1 | Identify Hepatitis C Patients | If a patient has a positive HCV RNA test | They are diagnosed with Hepatitis C | Patient(?p) ^ hasTestResult(?p, ?test) ^ HCVRNA_Test(?test) ^ PositiveResult(?test) -> HepatitisC_Patient(?p, yes) |
| 2 | Treatment Eligibility | If a Hepatitis C patient has a fibrosis stage between F0 and F2 | They are eligible for standard antiviral treatment | HepatitisC_Patient(?p) ^ hasFibrosisStage(?p, ?stage) ^ swrlb:greaterThanOrEqual(?stage, 0) ^ swrlb:lessThanOrEqual(?stage, 2) -> EligibleForStandardTreatment(?p, true) |
| 3 | Advanced Fibrosis or | If a Hepatitis C patient has a | They should be referred | HepatitisC_Patient(?p) ^ hasFibrosisStage(?p, ?stage) ^ |

| | | Cirrhosis Management | fibrosis stage F3 or F4 | for specialized management | swrlb:greaterThanOrEqual(?stage, 3) -> NeedsSpecializedManagement(?p, hospitalized) |
|---|---|---|---|---|---|
| 4 | | Monitoring | If a Hepatitis C patient is on antiviral treatment | They should be monitored every 4 weeks | HepatitisC_Patient(?p) ^ OnAntiviralTreatment(?p) -> needsMonitoring(?p, every4Weeks) |
| 5 | | Post-Treatment Follow-Up | If a Hepatitis C patient has completed antiviral treatment | They should have an HCV RNA test at 12 weeks post-treatment to confirm sustained virological response (SVR) | HepatitisC_Patient(?p) ^ CompletedTreatment(?p) -> needsTest(?p, HCVRNA_Test, postTreatment12Weeks) |
| 6 | | Identifying Non-Responders | If a Hepatitis C patient has a positive HCV RNA test at 12 weeks post-treatment | They are considered a non-responder | HepatitisC_Patient(?p) ^ CompletedTreatment(?p) ^ hasTestResult(?p, ?test) ^ HCVRNA_Test(?test) ^ PositiveResult(?test) ^ postTreatment12Weeks(?test) -> NonResponder(?p) |
| 7 | | Lifestyle Recommendations | If a patient is diagnosed with Hepatitis C | They should be advised to avoid alcohol and be vaccinated against Hepatitis A and B | HepatitisC_Patient(?p) -> needsLifestyleChange(?p, AvoidAlcohol) ^ needsVaccination(?p, HepatitisA) ^ needsVaccination(?p, HepatitisB) |
| 8 | | Identify Cirrhosis Patients | If a patient has liver biopsy results | They are diagnosed with | Patient(?p) ^ hasLiverBiopsyResult(?p, ?result) ^ CirrhosisStage(?result, F4) -> |

| | | indicating cirrhosis (F4 stage) | cirrhosis | Cirrhosis_Patient(?p) |
|---|---|---|---|---|
| 9 | Screening for Hepatocellular Carcinoma (HCC) | If a patient is diagnosed with cirrhosis | They should undergo ultrasound screening every 6 months | Cirrhosis_Patient(?p) -> needsScreening(?p, Ultrasound, every6Months) |
| 10 | Varices Screening | If a patient is diagnosed with cirrhosis | It should have an upper endoscopy to screen for varices | Cirrhosis_Patient(?p) -> needsScreening(?p, UpperEndoscopy) |
| 11 | Monitoring for Hepatic Encephalopathy | If a patient is diagnosed with cirrhosis | It should be monitored for signs of hepatic encephalopathy | Cirrhosis_Patient(?p) -> needsMonitoring(?p, HepaticEncephalopathySigns) |
| 12 | Ascites Management | If a patient with cirrhosis develops ascites | They should be treated with diuretics and sodium restriction | Cirrhosis_Patient(?p) ^ hasAscites(?p) -> needsTreatment(?p, Diuretics) ^ needsDietaryChange(?p, SodiumRestriction) |
| 13 | Referral for Liver Transplant Evaluation | If a patient with cirrhosis has decompensated liver disease | It should be referred for liver transplant evaluation | Cirrhosis_Patient(?p) ^ hasDecompensatedLiverDisease(?p) -> needsReferral(?p, LiverTransplantEvaluation) |
| 14 | Alcohol Abstinence | If a patient is diagnosed with cirrhosis | It should be advised to abstain from alcohol completely | Cirrhosis_Patient(?p) -> needsLifestyleChange(?p, AbstainFromAlcohol) |

### 3.10.1 Algorithm for the Hepatitis C Patient Treatment

Individuals diagnosed with a hepatitis C virus infection (viremia positive) require treatment. The treatment duration is influenced by several factors, including whether the patient has cirrhosis or non-cirrhosis, the presence of decompensation (such as ascites, variceal bleeding, hepatic encephalopathy, or infections), and whether the patient is treatment-naive or has previously undergone treatment with pegylated interferon (peg IFN) or direct-acting antivirals (DAAs). In India, DAAs are recommended as the first-line treatment. The specific combination of DAAs and the treatment duration depend on the presence or absence of cirrhosis and the genotype of the virus. The following algorithm guides the selection of the appropriate regimen and duration for treatment-naive hepatitis C patients, as shown in Figure 11 and algorithm converted in SWRL rules and it's implementation as shown in Figure 17.

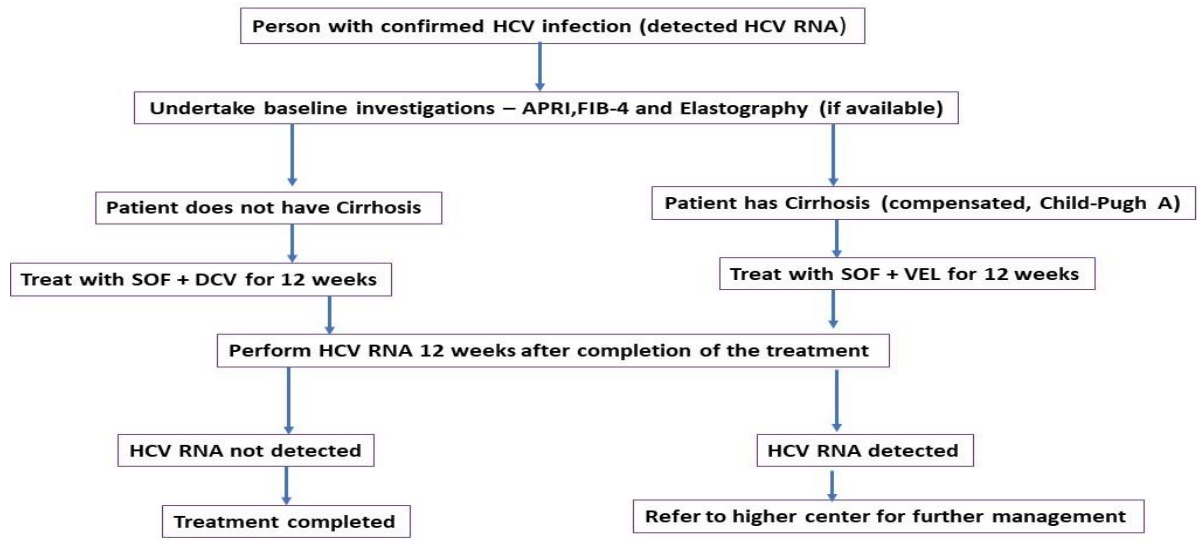

Figure 11: Algorithm of hepatitis C treatment

The patients who did not have a cirrhosis regime recommended (Sofosbuvir(400mg) and Daclatasvir(60mg)) for 12 weeks. Patients with cirrhosis compensated (Child-Pugh A) regime recommended (Sofosbuvir(400mg) + Velpatasvir(100mg)) for 12 weeks. Patients have cirrhosis-decompensated (Child-Pugh B and C) Regime recommended (Sofosbuvir(400mg) + Velpatasvir (100mg) & Ribavirin(600-1200mg)) for 12 weeks. This Liver ontology covers all this medication knowledge related to liver diseases. Medication is recommended based on rules and results that have taken the medication name for Indian scenarios-based on NVHCP guidelines.

### 3.11 SPARQL Query Results

SPARQL is a query language for querying and manipulating data stored in RDF format. It enables querying and searching for specific patterns, relationships, and information within RDF graphs, making it a fundamental tool for the semantic web. It provides a standardized way to retrieve and manipulate RDF data using a syntax similar to Structured Query Language (SQL). SPARQL allows users to make complicated searches using subject-predicate-object triples, and it supports a variety of query forms, including SELECT for result sets, CONSTRUCT for generating new RDF graphs, ASK for boolean results, and DESCRIBE for extracting metadata details. The syntax of SPARQL is adaptable, and it can manage remote and diverse data sources, allowing for easy integration and reasoning over linked data.

### 3.11.1 Real-Time Query using Apache Jena

Apache Jena is an open-source Java framework designed to develop applications related to the semantic web and linked data. This software framework offers a comprehensive set of tools and libraries to facilitate the manipulation and analysis of semantic data, namely in RDF, OWL, and SPARQL [28] [31]. In this work, the first step is to process the RDF data in batches. Figure 12 shows that batch number 62 has been processed in 8 ms, and then a SPARQL query is applied to these batches, and the output matches the filters. According to the rules extracted from the decision tree, it also functions effectively as it indicates that the Patient Category is 3, which signifies that the patient is suffering from fibrosis.

Figure 12: SPARQL Query and its Results based on batch processing using Apache Jena

### 3.11.2 SPARQL Query using Protégé

In Figure 13, it is observed that for the same SPARQL query that was used for Apache Jena, the output is the same as in Protégé. A framework view of the SPARQL Query tab from the Protégé tool shows all the fibrosis patients. It also validates the constructed SWRL rule and signifies that the proposed model is perfectly compatible with static and batch-processing data.

Figure 13: SPARQL Query and its result in Protégé

Table 9 shows the query which is used in Figure 14.

Table 9: SPARQL Query

```
PREFIX ns1: <http://schema.org/>
PREFIX rdf: <http://www.w3.org/1999/02/22-rdf-syntax-ns#>
PREFIX xsd: <http://www.w3.org/2001/XMLSchema#>
SELECT ?SNo ?ALT ?AST ?GGT ?ALB ?ALP ?BIL ?Category
WHERE {
    ?record rdf:type ns1:MedicalRecord ;
        ns1:SNo ?SNo ;
        ns1:ALT ?ALT ;
        ns1:AST ?AST ;
        ns1:GGT ?GGT ;
        ns1:ALB ?ALB ;
        ns1:ALP ?ALP ;
        ns1:Sex ?Sex ;
        ns1:BIL ?BIL ;
        ns1:Category ?Category.
    FILTER (?ALB >= 30.0)
    FILTER (?GGT >= 60.0)
}
ORDER BY ASC(?BIL)
```

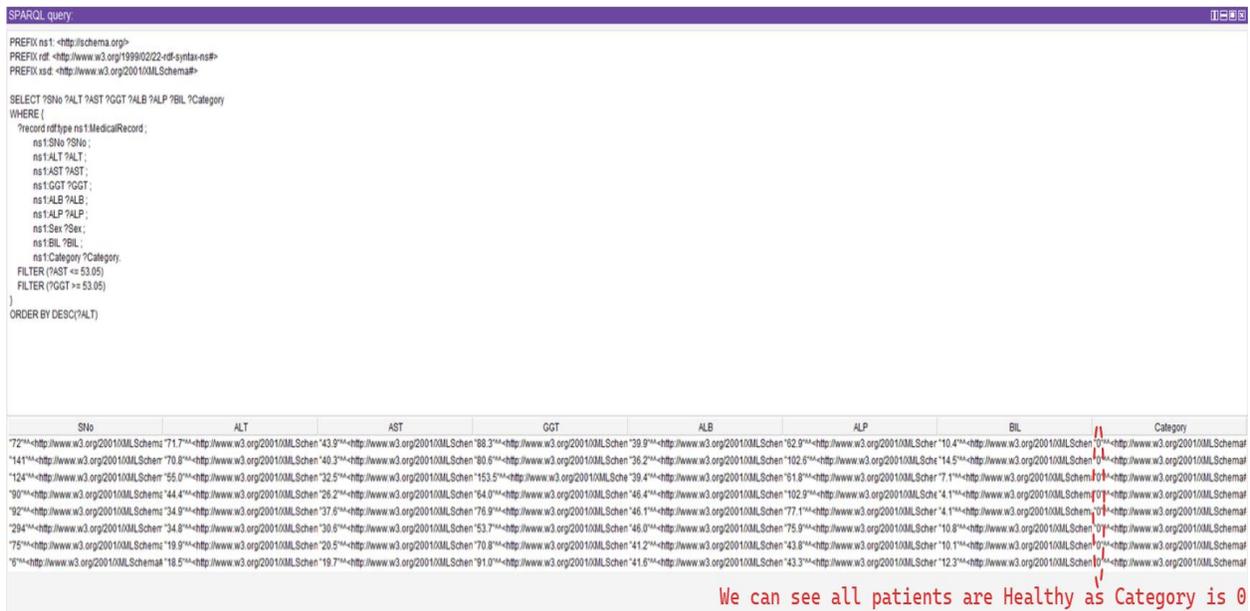
Figure 14: SPARQL query for Patients who come under category 0

## 3.12 Framework View of SWRL Rules

SWRL is a substantial OWL extension that is applicable for the development of rules inside a knowledge graph. SWRL may make logical inferences and deductions based on existing RDF data and ontology axioms. SWRL rules are expressed as logical expressions evaluated against RDF data to produce new information. The rules consist of two parts: an antecedent (also known as the "if" part) and a consequent (also known as the "then" part). The antecedent specifies the conditions that must be met for the rule to be applied, while the consequent specifies the actions to be taken if the rule is applied. Figure 16 shows the framework view of the SWRL rules and their results, with an explanation.

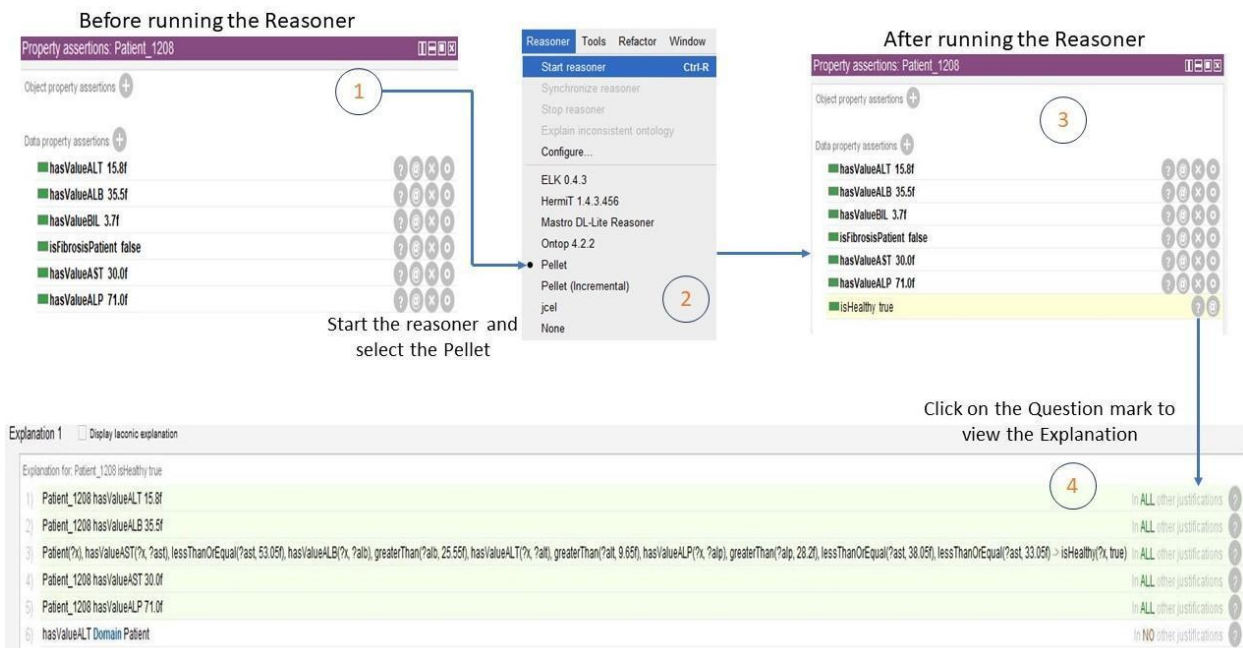

Figure 15: SWRL rules working process in Protege

The Knowledge Graph also gets validated for inconsistencies when Pellet Reasoner is started. In Figure 15, the details of patient 1208 can be visualized in step 1. After running the Pellet Reasoner in Step 2, the output of the SWRL Rule is obtained in yellow in Step 3, from which the explanation can be seen in Step 4. The satisfied SWRL rule gives the result that Patient 1208 is Healthy. The rules that are used for this are shown in Table 11.

Table 11: Rule that Patient 1208 satisfies is healthy

| Patient(?x) ^ hasValueAST(?x, ?ast) ^ swrlb:lessThanOrEqual(?ast, "53.05"^^xsd:float) ^ hasValueALB(?x, ?alb) ^ swrlb:greaterThan(?alb, "25.55"^^xsd:float) ^ hasValueALT(?x, ?alt) ^ swrlb:greaterThan(?alt, "9.65"^^xsd:float) ^ hasValueALP(?x, ?alp) ^ swrlb:greaterThan(?alp, "28.2"^^xsd:float) -> isHealthy(?x, true) |
|---|

### 3.12.1 XAI-based Explanation and Suggestion

Our study employed the OpenAI API, a sophisticated language model, to generate user-friendly explanations. We fed the results from SWRL reasoning into the OpenAI API, as depicted in Figure 16. This process allowed us to produce clear, understandable explanations of the dengue diagnosis, enhancing the transparency and interpretability of the system. Initially, we obtained results through SWRL reasoning and later used OpenAI API's NLP capabilities to convert these results into accessible explanations. By leveraging ChatGPT's ability to generate human-like responses, we made the complex reasoning process more comprehensible for users. This method effectively bridged the gap between intricate reasoning and user understanding, making the

SWRL outcomes more transparent and interpretable. We passed the SWRL output, including derived facts or conclusions, to the OpenAI API, which then generated concise, natural language explanations customized to the specific reasoning results [32].

In conclusion, while ontologies provide limited, localized explanations, the use of OpenAI API significantly enhanced user comprehension in the Liver ontology context. This approach improved users' understanding of the findings and encouraged interactive exploration and analysis of the results.

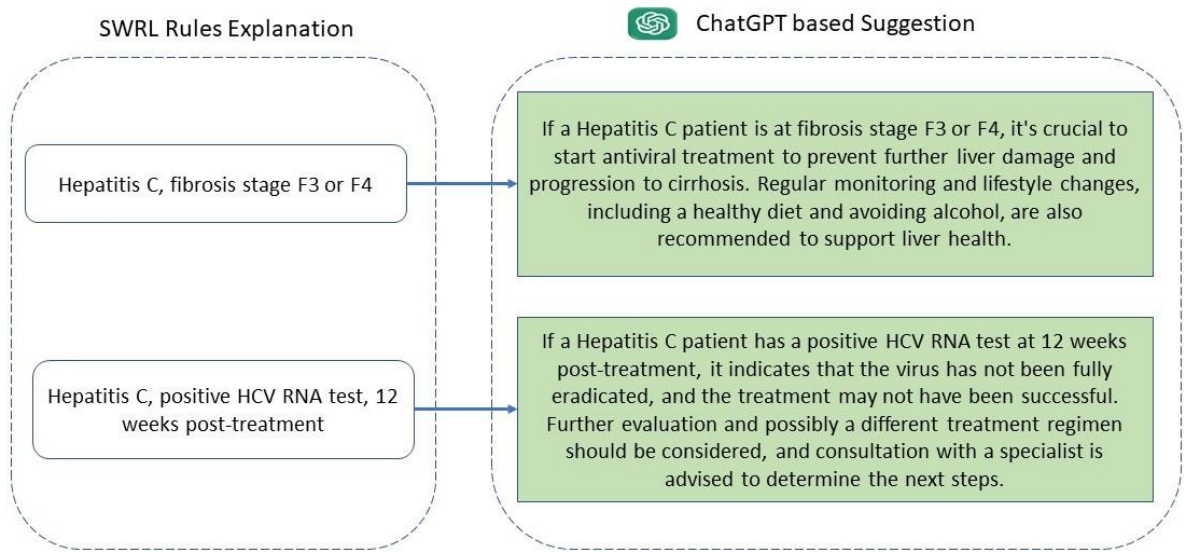

Figure 16: User Level suggestion based on SWRL rules

Figure 17 shows the use case scenario of how a patient is diagnosed with cirrhosis based on SWRL rules and how these rules prescribe medication based on the patient's medical tests. It also includes an explanation using XAI for user-level understanding, providing a concise and precise explanation with relevant suggestions. When a patient's medical values fall under the parameters of Rule 1, it indicates the presence of Cirrhosis. In such cases, the patient must undergo certain tests suggested by XAI to determine the class type of cirrhosis. If the test results categorize the condition as Class A (Child Pugh A), it is considered mild. According to Rule 4, appropriate medications are prescribed.

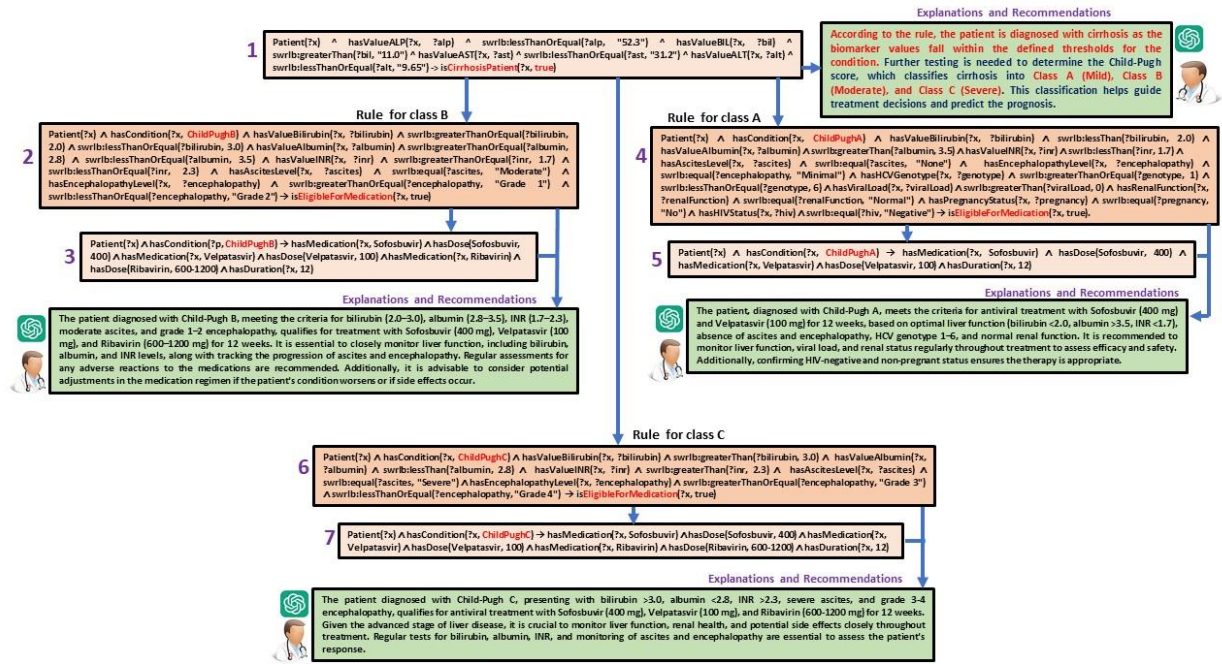

Figure 17: Use case scenario of how a patient with cirrhosis is diagnosed

Rule 5 specifies which medications are needed and the duration of treatment. XAI provides a detailed explanation based on Rules 4 and 5, including recommendations and additional information such as the need for regular liver function monitoring, confirmation that HIV is negative, and that the patient is not pregnant for ongoing medication safety.

If the patient falls under Class B (Child Pugh B), the situation is moderate and medications are prescribed based on Rule 2. The specific medications and their duration are outlined in Rule 5, and adjustments for any reactions during regular medication are described in Rule 3. XAI provides explanations and recommendations based on Rules 4 and 5, including possible adjustments in the medication regimen if adverse reactions occur.

For patients classified under Class C (Child Pugh C), which indicates a severe condition, medications are prescribed according to Rule 6. Rule 7 specifies both the medications and their duration. XAI offers detailed explanations and recommendations based on Rules 6 and 7, emphasizing the importance of regular tests for bilirubin and albumin and monitoring ascites and encephalopathy to assess the patient's response.

Based on the rules, patients are prescribed checkups for further classification of diseases. When the patient returns with the test reports from these checkups, OCR accommodates the reports and adds the details based on the liver disease ontology. Subsequently, rules are applied for further treatment, along with suggestions and recommendations from XAI.

## 4. Experimental Results

Numerous parameters, such as event processing time, rule deployment time, accuracy, precision, recall, F1-score, and SPARQL query performance time, are used in this section to evaluate the proposed ontology.

### 4.1 Performance of Event Processing Systems

Table 12 displays five batches, each with batch size 10, indicating that, despite injecting many rules, only some parsed the condition. For example, in batch 3, 5 rules parsed the condition, whereas in other batches, only 3 rules parsed. The variation in time occurs batch by batch because some rules are simple while others are complex, leading to time variations in milliseconds.

Table 12: Computation of event processing time

| Batch No. | Batch Size | No. of Rules that are parsed for event detection | Time in milliseconds |
|---|---|---|---|
| Batch 1 | 10 | 3 | 8 |
| Batch 2 | 10 | 3 | 2 |
| Batch 3 | 10 | 5 | 14 |
| Batch 4 | 10 | 3 | 9 |
| Batch 5 | 10 | 3 | 3 |

Table 13 shows the time (in milliseconds) it takes to process 5 rules for different batch sizes (20, 40, 60, 80, and 100). The data is recorded over 5 separate runs (1 to 5). As the batch size increases, the processing time generally increases as well, meaning larger batches take more time to process the same number of rules. For instance, with a batch size of 100, the processing time is consistently higher compared to smaller batch sizes like 20.

However, the times vary slightly between different runs, indicating that other factors, such as system performance or rule complexity, might also influence the processing time. This shows a trend where larger batches tend to take more time, but the relationship isn't always perfectly consistent as shown in Figure 18.

Table 13: Comparative analysis of rules processed across different times (in milliseconds) on different batch sizes

| Batches | Batch Size 20 (5 Rules) | Batch Size 40 (5 Rules) | Batch Size 60 (5 Rules) | Batch Size 80 (5 Rules) | Batch Size 100 (5 Rules) |
|---|---|---|---|---|---|
| 1 | 21 | 35 | 48 | 63 | 79 |
| 2 | 29 | 33 | 50 | 61 | 75 |
| 3 | 17 | 34 | 46 | 80 | 73 |
| 4 | 22 | 31 | 50 | 66 | 69 |
| 5 | 21 | 28 | 49 | 63 | 88 |

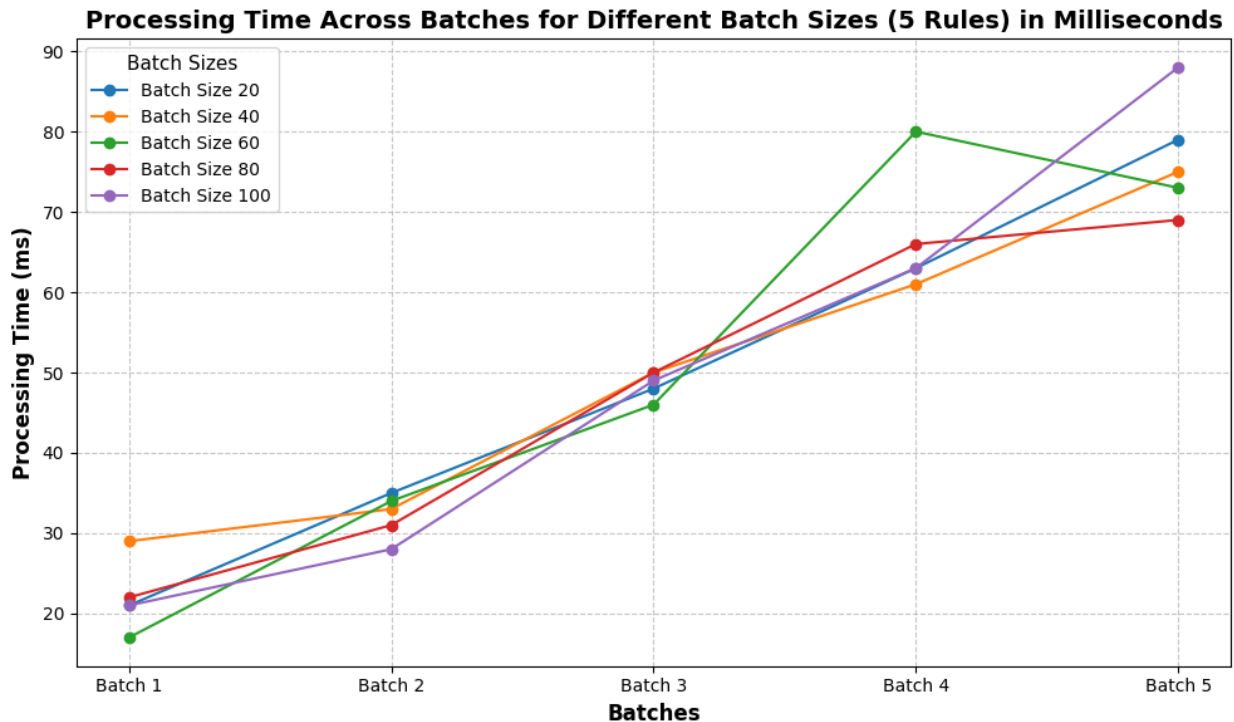

Figure 18: This graph shows the comparative analysis of rule processing time across different batch sizes

Table 14 presents the processing time (in milliseconds) for different rule sizes (4, 6, 8, 10, and 12 rules) with the same batch size of 50, measured over 5 runs. As the number of rules in each batch increases, the processing time generally increases as well. For example, with 4 rules, the processing time ranges from 22 to 26 ms, while with 12 rules, it increases to between 106 and 109 ms. This trend shows that larger rule sizes take longer to process. However, there are some variations in processing time between runs, suggesting that other factors such as system performance or the complexity of the rules may also influence the results as shown in Figure 19. Overall, the table highlights that more rules within a batch size of 50 lead to increased processing time.

Table 14: Comparative analysis of different rule sizes processed across different times (in milliseconds) on the same batch size

| Batches | Batch Size 50 (4 Rules) | Batch Size 50 (6 Rules) | Batch Size 50 (8 Rules) | Batch Size 50 (10 Rules) | Batch Size 50 (12 Rules) |
|---|---|---|---|---|---|
| 1 | 26 | 52 | 72 | 91 | 106 |
| 2 | 24 | 52 | 74 | 92 | 109 |
| 3 | 24 | 56 | 72 | 91 | 108 |
| 4 | 22 | 23 | 81 | 96 | 108 |
| 5 | 26 | 52 | 75 | 90 | 109 |

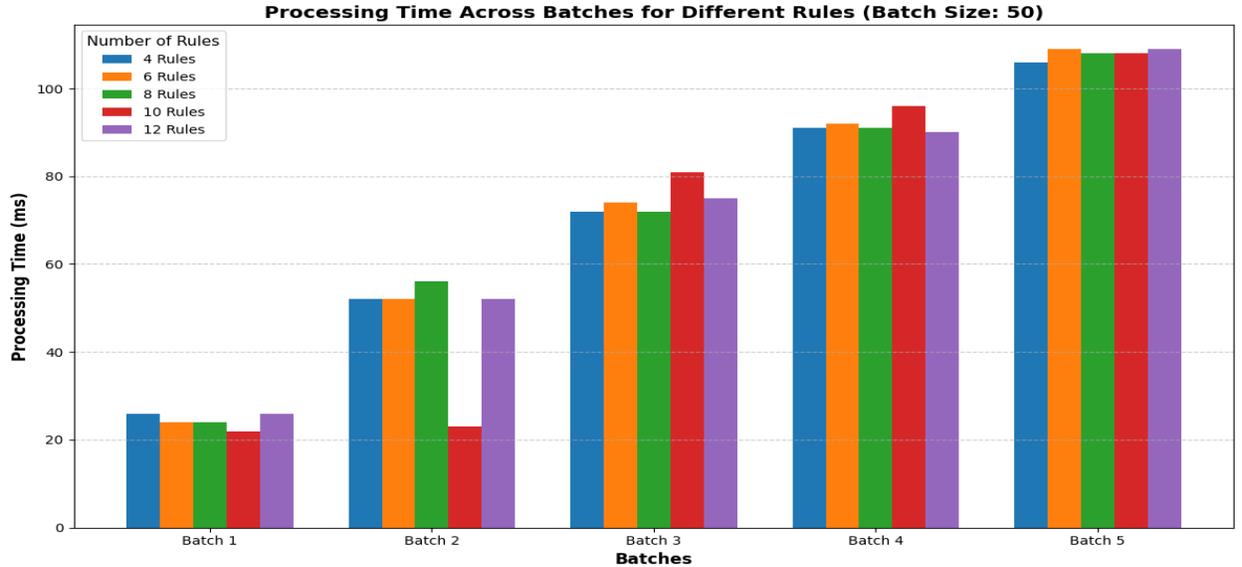

Figure 19: This graph compares rule processing times for different rule sizes on the same batch size

## 4.2 Rule Deployment Time

The time needed to deploy the rule is measured in the Apache Jena processing engine by keeping track of changes (including adding, updating, and removing) made to rules while running. The experiment results for two rules of various events are presented as a graph in Figure 20. There is a positive correlation between the number of events and the time required for deployment.

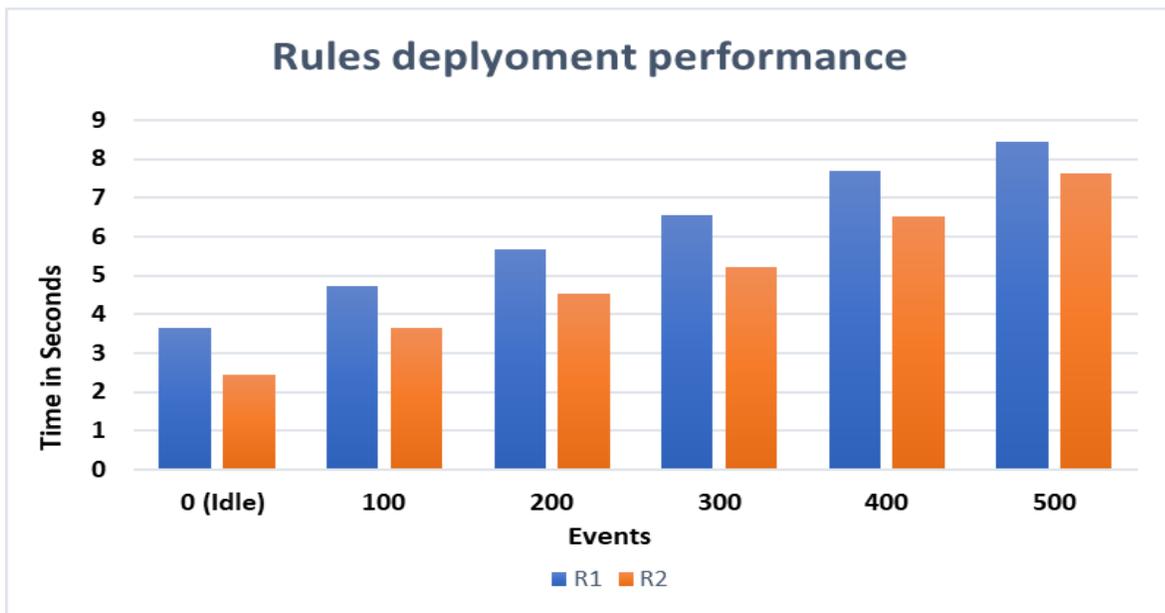

Figure 20: Graph of Rule Deployment times on different events

## 4.3 Calculation of Precision, Recall and F1 Score

In the context of decision trees, accuracy, F-score, precision, and recall are used as assessment metrics to analyze the model's performance in classification tasks. Since the difference between entropy and Gini is insignificant, we chose Gini as the evaluation criterion for the decision tree model in this work, and the results are shown in Table 15.

Table 15: Decision Tree Metrics

| Metrics | Gini Criterion | Entropy Criterion |
|---|---|---|
| Average Accuracy | 93.31 % | 93.02 % |
| Average Precision | 92.73 % | 92.35 % |

| Average Recall | 93.31 % | 93.02 % |
|---|---|---|
| Average F-Measure | 91.72 % | 91.76 % |

### 4.4 SPARQL query performance time based on different RDF data

Table 16 displays the query's performance in terms of time. Each of the five questions exhibits varying levels of complexity, indicating that the query differs based on the specific occurrence. The execution of Query 4 takes longer due to its high level of complexity.

Table 16: Shows the query execution duration in seconds across various RDF datasets

| Query | RDF Data 1 | RDF Data 2 | RDF Data 3 | RDF Data 4 |
|---|---|---|---|---|
| Q1 | 5 | 7 | 14 | 26 |
| Q2 | 7 | 11 | 16 | 30 |
| Q3 | 6 | 15 | 24 | 33 |
| Q4* | 12 | 30 | 21 | 63 |
| Q5 | 7 | 15 | 21 | 43 |

### 4.5 Ontology Metrics-based Evaluation

Table 17 shows the count of different metric values to determine Attribute Richness (AR), Class Richness (CR), Average Population (AP), and Relationship Richness (RR) to prove the effectiveness of the proposed ontology.

Table 17: Metrics count of the designed ontology

| Metrics | Count |
|---|---|
| Axiom | 8263 |
| Logical axiom count | 2983 |
| Declaration axioms count | 1252 |
| Class count | 125 |
| Object property count | 27 |
| Data property count | 28 |
| Individual count | 615 |
| Annotation Property count | 21 |
| SubClassOf | 12 |
| DisjointClasses | 96 |
| SubObjectPropertyOf | 221 |
| ObjectPropertyDomain | 156 |

**Schema metrics:** The ontology can alternatively be described using a 5-tuple model, denoted as O = <C, Dr, Sc, Re, Ind>, where C represents classes, Dr denotes data properties (attributes), Sc indicates subclasses, Ind stands for individuals, and Re signifies relations between classes. Metrics can be assessed based on AR, RR, CR, and AP.

RR measures the depth of connections between concepts in an ontology. It is calculated using Equation 1:

$$RR = \frac{|Prop|}{|Sub\ class| + |Prop|} \quad \ldots\ldots\ldots\ldots\ldots\ldots\ldots\ldots\ldots\ldots\ldots\ldots\ldots\ldots(1)$$

where |Prop| is the total number of properties, including attribute data and object characteristics (class relationships).

AR is calculated by averaging the number of attributes over the entire class, as shown in Equation 2:

$$AR = \frac{|Attribute|}{|Class|} \quad \ldots\ldots\ldots\ldots\ldots\ldots\ldots\ldots\ldots\ldots\ldots\ldots\ldots\ldots(2)$$

where |attribute| represents the total number of data attributes.

CR indicates the amount of real-world knowledge conveyed through the ontology. It is calculated with equation 3 by dividing the number of classes with instances by the total number of classes:

$$CR = \frac{|Class\ with\_instance|}{|Class|} \quad \ldots\ldots\ldots\ldots\ldots\ldots\ldots\ldots\ldots\ldots\ldots\ldots\ldots\ldots(3)$$

AP determines the average number of individuals in each class, expressed in equation 4:

$$AP = \frac{|Individual|}{|Class|} \quad \ldots\ldots\ldots\ldots\ldots\ldots\ldots\ldots\ldots\ldots\ldots\ldots\ldots\ldots(4)$$

The computed values of different evaluation metrics for the newly designed ontology are mentioned in Table 18.

Table 18: Different parameters-based metrics evaluation

| Ontology Metrics | Results |
|---|---|
| Attribute Richness | 0.58 |
| Class Richness | 0.56 |
| Average Population | 2.67 |
| Relationship Richness | 0.7 |

**4.6 Results from the evaluation of the OCR engine**

Table 19 illustrates how the OCR engine extracts information from patient test reports. It shows the number of words extracted and identifies which words are helpful based on the ground truth. We can perform a performance analysis using this data, as summarized in Table 16.

Table 19: Text extraction performance based on OCR

| Clinical Domain | Ground Truth | True Positive | False Negative | Precision | Recall |
|---|---|---|---|---|---|
| Cirrhosis | 675 | 595 | 80 | 0.882 | 0.884 |
| Fibrosis | 492 | 424 | 68 | 0.861 | 0.862 |
| Hepatitis C | 473 | 365 | 108 | 0.771 | 0.771 |
|  | 1640 | 1384 | 256 | 0.844 | 0.844 |

**5. Discussion**

In this work, raw data undergoes pre-processing and is converted into RDF format. The RDF data is then batch-processed using Apache Jena. DT are used to generate rules based on the dataset, which are validated against ground truth parameters. These DT rules are subsequently applied for event detection in batch-processing mode. An ontology is developed using BFO, PCD, and the RDF data. To operationalize the ontology, DT rules and government medical guidelines are converted into SWRL rules and integrated into the ontology using the Protégé tool.

Event detection is performed by executing SPARQL queries on batch-processed data and ontology. The query results generate insights into a patient's liver health, such as confirming a healthy liver or diagnosing potential liver diseases. If symptoms of liver disease are detected, recommendations for medical tests and precautionary measures are provided. Once the patient completes the recommended medical tests, the test reports are processed using OCR and NLP techniques to extract relevant information [33][34]. This extracted information is mapped to the PCD ontology, which organizes it based on the domain. A similar process is applied to the Jena and Liver Diseases ontologies. If positive symptoms are detected, appropriate medications and suggestions are provided as required. This complete model functions as a comprehensive decision support system for diagnosing and treating liver diseases.

Event processing times vary due to the sequential condition checks required for some rules. These variations depend on the complexity of the conditions and the rules being evaluated. Two scenarios were evaluated during testing:

1. Without event processing: The processing engine remains inactive or handles zero events.
2. With event processing: The system handles events at varying rates, from 0 to 500 events per second.

To evaluate average rule deployment time, 20 tests were conducted for each scenario. The tests incrementally increased the number of rules updated with each iteration.

A set of five SPARQL queries was executed on different RDF datasets. Query execution time varied based on the dataset size and the complexity of the queries. Ontology metrics were used to assess the performance of the ontologies, providing quantitative insights into their structure, content, and usability. The evaluation incorporated metrics such as event count and deployment time in batch-processed data, offering valuable insights into the ontology's scalability and effectiveness.

The knowledge graph is designed to efficiently analyze and diagnose liver diseases using a dataset of 615 samples, covering hepatitis C, fibrosis, cirrhosis, and cases with symptoms but no confirmed disease. By integrating NVHCP guidelines, it can handle complex diagnostic scenarios. We specifically designed the knowledge graph to diagnose liver disease.

Overall, this model represents a significant advancement in liver disease diagnosis and ontology-based decision support systems. It addresses previous limitations and introduces innovative techniques to enhance efficiency and performance. Table 20 provides a comparative analysis of existing work in this domain.

Table 20: Comparison with existing work

| Reference | Objective | Methodology | Technology used | Result |
|---|---|---|---|---|
| Yunzhi et al. [17] | Construct hepatitis ontology. | Utilize ontology principles for clarity, consistency, and scalability; Define structure, extract concepts, and hierarchy; Enhance semantic search. | Protégé 3.4.1 is for ontology editor; Jena environment is for semantic web tool. | Improved query accuracy and retrieval precision. |
| Banihashem et al. [8] | Develop a predictive model for fatty liver disease using data mining and ontology to share knowledge with clinical systems. | Collected clinical data, preprocessed it, applied decision tree classification, created an ontology model, and used SWRL rules for inference. | RapidMiner Studio, Protégé 5, Python, Drools. | Achieved 73.79% accuracy, aligned ontology detection with decision tree, and reviewed accuracies of related methods ranging from 61.4% to 87.48%. |
| Anwaar et al. [14] | Develop a liver hepatitis ontology for intelligent healthcare systems and physicians' reference. | Ontology development, gathering information from medical experts, designing ontology hierarchy, and validating with medical professionals. | Protégé for ontology editing and implementation, WEB-V-OWL for visualization. | A comprehensive ontology covering hepatitis A, B, C, and D, with detailed information on causes, symptoms, treatments, and diagnostic methods. |

| Proposed model | Developed predictive models for liver diseases using a Combination of BFO, DT rules, Apache Jena, and SPARQL query. | Collect clinical information for the development of ontology based on BFO, PCD ontology, OCR, NLP, SWRL rules-based DT, and medical guidelines; based on these rules, diseases are detected, based on batch processing and SPARQL applied based on the condition and ChatGPT based suggestion. | Protege 5.5.0, Apache jena, OCR, NLP, Python and XAI. | Query-based, event-based, and Ontology metrics evaluation, DT, OCR-based, Precision, Recall, F-measure, and Accuracy based on Gini and Entropy. Rules development evaluation. |
|---|---|---|---|---|

## 6. Conclusion and Future Work

A knowledge-based liver disease ontology is developed using the upper-level ontology BFO. This ontology encompasses extensive information related to liver disease derived from government medical guidelines. Based on these guidelines, SWRL rules are created, with some rules adapted from DT rules, to aid in decision-making related to liver disease. This method employs SPARQL queries on the ontology, followed by batch processing, which yields additional information about individuals. Apache Jena is used to stream data in batch processing, using the developed rules for event detection and queries based on these events. This approach offers numerous benefits and has the potential to function effectively in real-time scenarios.

The developed models incorporate rules derived from specific datasets and universal medical guidelines, making them precise and broadly applicable. These rules assist in generating explainable suggestions using ChatGPT at the user level. Additionally, OCR is used to extract information from patient medical reports, and NLP techniques are employed to process this information and convert it into the appropriate format for integration into the PCD ontology and subsequently into the Liver ontology. This process completes the model for diagnosing and treating liver diseases. The limitation of this work is the need for adjustments during model deployment to ensure alignment with the rules while preserving the integrity of other features. Additionally, the dataset used contained 615 samples diagnosed by the current model, but more diverse data is needed to further enhance the model's capability and reliability.

We will enhance the model in future work by incorporating diverse liver disease datasets and developing advanced rule-based reasoning for a heavy weight ontology. We will use big data technologies like Hadoop and Spark to enhance scalability, efficiency, and real-time analytics, thereby ensuring broader applicability and robustness in liver disease diagnosis.

**Acknowledgment**

This research is supported by the "Extra Mural Research (EMR) Government of India Fund by Council of Scientific & Industrial Research (CSIR)," Sanction letter no. – 60(0120)/19/EMR-II.

**Declaration**

**Competing interests**

The authors have no competing interests to declare relevant to this work's content.

**Author's contribution statement**

**Ritesh Chandra:** Conceptualization, Data curation, Methodology, Visualization, Writing – original draft. **Satyam Rastogi:** Formal analysis, Methodology, Software. **Sonali Agarwal:** Formal analysis, Investigation, Supervision, Writing – review & editing. **Sadhan Tiwari:** Writing & editing.

**Data availability and access**

The study's data was taken from a website and is freely accessible.

**Ethics approval and consent to participate**

Not Applicable.

**Funding and Acknowledgment**

The authors thank the Ministry of Education and the Indian Institute of Information Technology, Allahabad, for providing the necessary materials to complete this work.

**Conflicts of interests**

All authors declare no conflicts of interest in the presented work.

**References**:-